\documentclass[conference]{IEEEtran}
 
\usepackage{cite}
\usepackage{amsmath,amssymb,amsfonts}
\usepackage[ruled,vlined,linesnumbered]{algorithm2e}
\usepackage{array}
\usepackage{graphicx}
\usepackage{textcomp}
\usepackage{booktabs}
\usepackage{longtable}
\usepackage{subcaption}
\usepackage[table]{xcolor}
\usepackage{pgf-pie}
\usepackage{tabularx}

\usepackage[no-math]{fontspec}
\usepackage{polyglossia}

\setmainlanguage{english}
\setotherlanguage{russian}

\defaultfontfeatures{Ligatures=TeX}

\newfontfamily\russianfont{Tempora}[
    Script=Cyrillic,
    Extension=.otf,
    UprightFont=*-Regular,
    ItalicFont=*-Italic,
    BoldFont=*-Bold,
    BoldItalicFont=*-BoldItalic
]

\newfontfamily\kazakhfont{Tempora}[
    Script=Cyrillic,
    Extension=.otf,
    UprightFont=*-Regular,
    ItalicFont=*-Italic,
    BoldFont=*-Bold,
    BoldItalicFont=*-BoldItalic
]

\begin{document}
%

\bibliographystyle{IEEEtran}


\title{Not All Color Categories Are Equally Stable: A Multilingual Free Color Naming Experiment}






\author{
\IEEEauthorblockN{Nuray Toganas, Adilet Yerkin, Elnara Kadyrgali, Muragul Muratbekova, Aron Karatayev, Pakizar Shamoi\IEEEauthorrefmark{1}}
\IEEEauthorblockA{School of Information Technology and Engineering,\\
Kazakh-British Technical University,
Almaty, Kazakhstan 050000\\ 
Email: p.shamoi@kbtu.kz
}
}



\maketitle

\begin{abstract}

Color naming is an important part of human color perception. Its task is to allow people to describe continuous colors using discrete color categories. However, the boundaries between color categories are often unclear, and some colors may be perceived differently depending on their saturation and brightness. While certain color categories remain recognizable across a wide range of shades, others may be associated with different color names when their appearance changes. This study investigates the consistency of color naming for red, yellow, and green color categories using a free color-naming experiment. A set of 18 color samples was selected from the COLIBRI dataset to represent different shades of these colors. Participants (n = 92) were asked to freely assign color names to each sample in Kazakh, Russian, or English without being limited to predefined categories. The results show that color categories differ in their consistency. Green shades were consistently identified as green despite variations in appearance, whereas yellow shades received a wider variety of names, including gold- and brown-related descriptions. Red shades showed moderate naming consistency. Our findings suggest that some color categories occupy broader perceptual regions than others and may therefore be more robust to visual variations. The study results can be used to develop perceptually meaningful color models and color naming systems.

\end{abstract}

\IEEEpeerreviewmaketitle

\section{Introduction}

Are they all green? A matcha, a traffic signal, an unripe fruit, and a crocodile would all be described as green by most observers. Yet these stimuli differ substantially in their visual appearance, raising an important question: how much variation can a color undergo while still being perceived as belonging to the same category?


Color naming is a fundamental aspect of human color perception. It serves as a bridge between continuous color stimuli and discrete linguistic categories \cite{Twomey2021}. However, the boundaries of these categories are vague \cite{shamoi2025colibrifuzzymodelcolor, Chamorro-Martínez2017Fuzzy}, and different color regions may show varying levels of naming stability. While some color categories maintain their identity across a broad range of shades, others appear more sensitive to changes in hue, saturation, and intensity \cite{Witzel2019Variation, Witzel2013Focal}.

This study addresses this gap through a free color-naming experiment involving multiple shades of \textit{red, yellow,} and \textit{green}. 
The visible color spectrum is a physical continuum, yet speakers of every language divide it into a finite set of named categories. 


In this study, we investigate the naming consistency of red, yellow, and green color categories using a free color-naming experiment, resulting in mapping from continuous perceptual space to discrete linguistic labels. A set of 18 color samples derived from the COLIBRI (Color Linguistic-Based Representation and Interpretation) \cite{shamoi2025colibrifuzzymodelcolor} dataset was selected to represent variations of these color categories.




The study has the following contributions:
\begin{itemize}
    \item We conduct a multilingual free color-naming experiment involving Kazakh, Russian, and English responses to investigate the perceptual stability of \textit{red, yellow}, and \textit{green }color categories.

    \item We analyze the consistency and diversity of color names assigned to different shades derived from the COLIBRI dataset and examine how naming stability varies across color categories.
\end{itemize}

The paper is structured as follows. This section is the Introduction. Section II provides an overview of the literature on color naming and categorization. Methodology, including the experiment, is presented in Section III.  Results and Discussion are presented in Sections IV and V, respectively. Finally, concluding remarks are drawn in Section VI.


\section{Related Work}

Color naming enables humans to transform a continuous color spectrum into a finite set of linguistic categories. One of the most influential theories in this field is the Basic Color Terms (BCT) theory \cite{berlin1969basic}, which argues that languages tend to develop a limited set of basic color categories in a predictable order. Subsequent studies have demonstrated that basic categories such as \textit{red, green, blue, yellow, black,} and \textit{white} form coherent, high-consensus categories and appear to have a special status in English and many languages \cite{mylonas2020, Lindsey2014The, Lindsey2006Universality}. 
Furthermore, multilingual variability in color naming reflects language-dependent differences in lexical access and semantic categorization strategies \cite{kay1997color, winawer2007color}.

Recent work has also explored the development of standardized color-naming systems based on  multisource datasets. The study \cite{sabitkyzy2026universalcolornamingsystem} proposed a clustering-based framework that aggregates over 19,000 color-name pairs from diverse sources and organizes them into perceptually meaningful 280 color categories.

\begin{figure*}[ht!]
    \centering
    \includegraphics[width = 0.95 \linewidth]{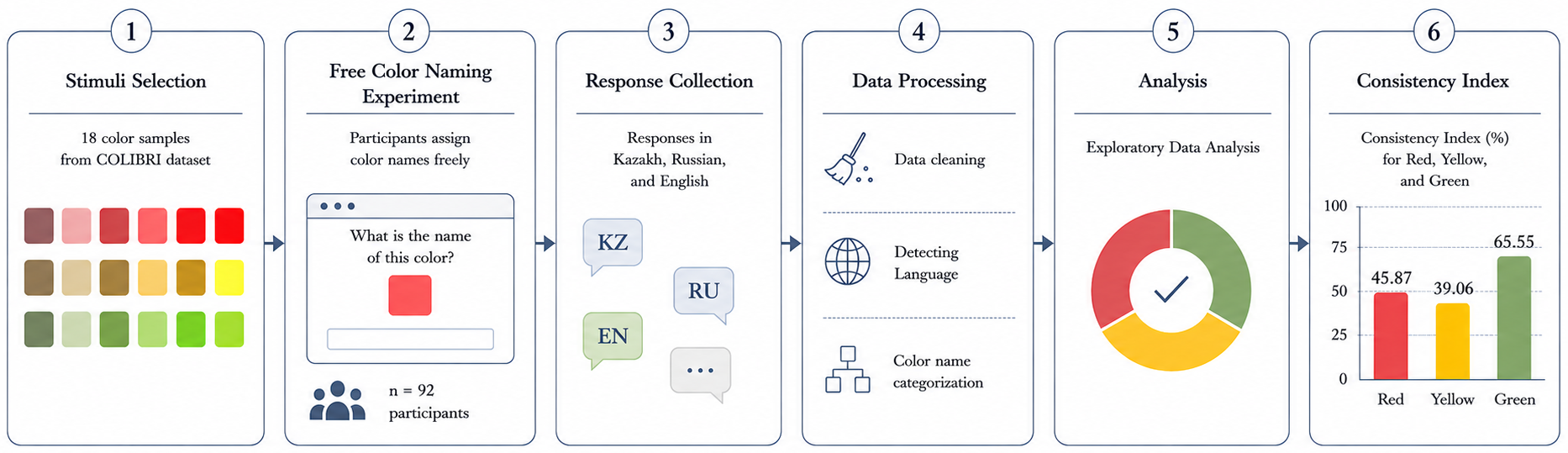}
    \caption{Experimental workflow for evaluating color category consistency using multilingual free color-naming responses}
    \label{fig:placeholder}
\end{figure*}

Although basic color categories are widely recognized, their boundaries are not sharply defined and often exhibit gradual transitions in perceptual color space \cite{witzel2018, regier2007, Shamoi14}. Modern theories describe color categories as regions organized around prototypical or focal colors, which represent the best examples of a category \cite{Regier2005Focal, Abbott2016Focal, Morimoto2022Invariant}. Colors located near these category centers tend to be named consistently, whereas colors near category boundaries frequently receive multiple alternative labels \cite{emery2017, mylonas2020}. This suggests that color naming is probabilistic rather than deterministic \cite{cibelli2016, regier2007}.

Recent studies suggest that the contradiction between continuous color perception and discrete linguistic categories can be explained through probabilistic models of category formation \cite{cibelli2016, witzel2018}. According to these models, color naming is organized around perceptual prototypes that are widely shared across populations, while category boundaries remain flexible and influenced by linguistic and cultural factors \cite{regier2007}. Consequently, color categories emerge from the interaction between universal perceptual mechanisms and language-specific conventions \cite{gong2019, shunbo2025}. Empirical studies further indicate that color naming is constrained by the neurophysiological properties of human vision \cite{abdou2021}. At the same time, categorical perception effects show that colors belonging to the same linguistic category are often perceived as more similar than physically equivalent color differences crossing category boundaries \cite{emery2017, forder2017}. These findings suggest that color categories are structured around stable perceptual centers, whereas their boundaries remain gradual and variable.

Previous research has shown that color categories are not equally distributed within perceptual color space. Some categories appear more robust to variations in hue, saturation, and brightness, while others are more sensitive to such changes. Focal colors, which represent the best examples of a category, tend to maintain their identity under changing viewing conditions \cite{Morimoto2022Invariant}. Across more than 110 languages, these best examples cluster near the prototypes of \textit{white, black, red, green, yellow, }and \textit{blue }\cite{Regier2005Focal, Abbott2016Focal}. Furthermore, changes in saturation and lightness have been shown to influence color naming and category membership \cite{emery2017, witzel2018}. While some hues preserve their categorical identity across a wide range of appearances, others are more likely to be associated with neighboring categories \cite{mylonas2020, Morimoto2022Invariant}. Together, these findings suggest that color categories may exhibit different levels of perceptual stability. However, the extent to which entire color categories preserve naming consistency under variations in appearance remains insufficiently understood.


Although many studies have examined color categorization, focal colors, and cross-linguistic color naming, limited attention has been paid to comparing the naming stability of different color categories.  Understanding these differences is important for perceptual color modeling and the development of human-centered color representation systems \cite{burambekova25}.

\begin{figure}[tb]
    \centering
    \includegraphics[width=0.6\linewidth]{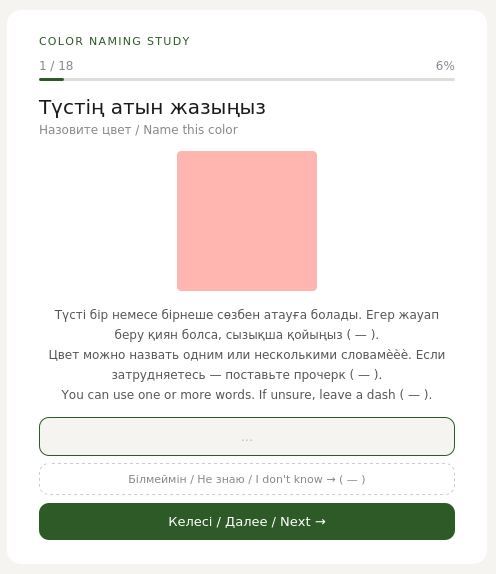}
    \caption{Sample from the experiment}
    \label{fig:experiment}
\end{figure}

\section{Methodology}

To investigate color consistency across different color categories, we conducted an online experiment during the conference on May 30, 2026.


We collected the information, including age and gender, 
as well as informed consent for the use of respondents' data in 
our research, via an online form (see Fig. \ref{fig:experiment}) distributed to students, staff, 
and guests of Kazakh-British Technical University. A total of 
92 participants participated (37 female, 54 male, 1 other; age 
range: 19--40 years). Participants were allowed to complete 
the form  in any language they were comfortable with.

\begin{table}[h!]
\centering
\caption{Color stimulus from COLIBRI Fuzzy Model}
\label{tab:color_descriptors}
\resizebox{0.9\columnwidth}{!}{%
\begin{tabular}{clllc}
\toprule
\textbf{ID} & \textbf{Hue} & \textbf{Saturation} & \textbf{Intensity} & \textbf{Color} \\
\midrule

4  & red    & low    & medium & \cellcolor[RGB]{142,95,95} \\ 
5  & red    & low    & light  & \cellcolor[RGB]{255,177,177} \\ 
7  & red    & medium & medium & \cellcolor[RGB]{204,64,64} \\ 
8  & red    & medium & light  & \cellcolor[RGB]{255,119,119} \\ 
10 & red    & high   & medium & \cellcolor[RGB]{255,14,14} \\ 
11 & red    & high   & light  & \cellcolor[RGB]{255,26,26} \\ 

22 & yellow & low    & medium & \cellcolor[RGB]{122,115,95} \\
23 & yellow & low    & light  & \cellcolor[RGB]{226,214,177} \\ 
25 & yellow & medium & medium & \cellcolor[RGB]{144,124,64} \\ 
26 & yellow & medium & light  & \cellcolor[RGB]{255,230,119} \\ 
28 & yellow & high   & medium & \cellcolor[RGB]{180,138,14} \\
29 & yellow & high   & light  & \cellcolor[RGB]{255,255,26} \\ 

31 & green  & low    & medium & \cellcolor[RGB]{108,128,95} \\
32 & green  & low    & light  & \cellcolor[RGB]{201,239,177} \\
34 & green  & medium & medium & \cellcolor[RGB]{103,164,64} \\
35 & green  & medium & light  & \cellcolor[RGB]{192,255,119} \\
37 & green  & high   & medium & \cellcolor[RGB]{96,222,14} \\
38 & green  & high   & light  & \cellcolor[RGB]{178,255,26} \\ 

\bottomrule
\end{tabular}
}
\end{table}

The stimuli (n=18) for the experiment were selected from the COLIBRI  dataset \cite{shamoi2025colibrifuzzymodelcolor}. COLIBRI is a human perception-based fuzzy color model that bridges the gap between computational color representations and human visual perception. Unlike traditional color models such as RGB, HSV, or CIE Lab, which rely on fixed numerical partitions, COLIBRI integrates fuzzy set theory with linguistically grounded color categories, enabling soft transitions between categories \cite{Color_model}.

The selected stimuli across three perceptual dimensions: Hue (\textit{red}, \textit{green}, \textit{yellow}), Saturation (\textit{low}, \textit{medium}, \textit{high}), and Intensity (\textit{medium}, \textit{light}), ensure uniform coverage of the color space (see Table \ref{tab:color_descriptors}). The general study workflow, including stimulus selection, response collection, data processing and analysis, and consistency calculation, is shown in Fig. \ref{fig:placeholder}.

For each of the 18 stimuli, participants were asked to name the color displayed in the box. They were allowed to use one or more words to describe the color. If they have difficulty, they could leave a dash or select the "I don't know" option.

We analyzed all data obtained from the experiment as presented in Algorithm \ref{algo:free_color_naming}. We began with a preprocessing step, in which unrelated responses, such as emojis and dashes, were removed, and text normalization was performed. A small proportion of responses (4.13\%) consisted of dashes or ``I don't know'' selections, indicating cases where participants were unable to assign a color name to the presented stimulus. These cases were excluded from the analysis. Responses in different languages were mapped to a unified color lexicon based on predefined color categories. This ensured consistency in semantic interpretation across multilingual inputs. Representative mappings are shown in Table~\ref{tab:color-category-dictionary}.


The resulting cleaned dataset was used for subsequent analysis and evaluation of color naming patterns. 

\begin{algorithm}[t]
\DontPrintSemicolon
\caption{Free Color Naming Consistency Analysis}
\label{algo:free_color_naming}

\KwIn{Dataset $D=\{(a_i,r_i)\}_{i=1}^{N}$, where $a_i$ is a color stimulus and $r_i$ is the corresponding free color naming response}
\KwOut{Consistency scores $\{C_h\}$ for all hue categories}

\ForEach{$(a_i,r_i)\in D$}{
    $(id_i,h_i,s_i,v_i)\leftarrow ExtractAttributes(a_i)$\;

    $\hat{r}_i \leftarrow Normalize(r_i)$\;

    $l_i \leftarrow DetectLanguage(\hat{r}_i)$\;

    $c_i \leftarrow ColorCategoryMap(\hat{r}_i)$\;

    $g_i \leftarrow SemanticMap(\hat{r}_i)$\;
}

\ForEach{hue category $h$}{
    $D_h \leftarrow \{i \mid h_i=h\}$\;
    $C_h \leftarrow \frac{1}{|D_h|}\sum_{i\in D_h}\mathbb{I}(c_i=h_i)$\;
}

Generate frequency tables for $\{c_i\}$\;
Generate word clouds for each hue category\;


\KwRet{$\{C_h\}$}
\end{algorithm}

\begin{table}[t]
\caption{Dictionary-based mapping of free color naming responses}
\label{tab:color-category-dictionary}
\centering
\scriptsize
\begin{tabularx}{\columnwidth}{
>{\raggedright\arraybackslash}p{0.12\columnwidth}
X
}
\toprule
\textbf{Category} & \textbf{Mapped response values} \\
\midrule
golden & golden, brown gold, золотой, темно-золотистый, грязно золотой... \\

red & red, brick red, burgundy red, scarlet red, neon red, muted red, грязно красный, қызыл, ашық қызыл, анық емес қызыл... \\

pink & pink, deep pink, pastel pink, pinkish, розовый, розовая пудра, телесно-розовый, красно-розовый, коричневый розовый, грязно розовый, ... \\

coral & coral, коралловый, светло-коралловый \\

peach & peach, персиковый \\

burgundy & бордовый, бургундский, пыльный бордовый, светлый бургунди... \\

yellow & yellow, lemon yellow, butter yellow, wheat yellow, кислотно-желтый, грязно-желтый, сары, сарғыш... \\

mustard & mustard, горчичный, темно-горчичный, бледно-горчичный \\

green & green, matcha green, lime green, olive green, кислотно-зеленый, темно-еловый зеленый, жасыл, травяной зеленый... \\

salad & lettuce, salad, салатовый, ядовито-салатовый, бледно-салатовый... \\

lime & lime, лайм, лаймовый, светло лаймовый \\

brown & brown, brownish, pinkish brown, grayish brown, коричневый, коричневая слива, қоңыр... \\

gray & gray, warm gray, серый, сұр, ржаво-серый, серый асфальт, коричнево-серый... \\

brick & brick, кирпичный \\

herbs & травяной, травянистый \\

maline & малиновый, грязно-малиновый, глиннино малиновый \\

lemon & лимонный, лимон, лимонный курд, цвет лимонной тарталетки \\

beige & beige, бежевый, темно бежевый, темно-бежевый, бежевый пастельный, бэйдж, светло бежевый \\

sand & песочный, бледно песочный \\

ohra & охра, бледная охра, грязная охра, светлая охра \\

olive & грязно-оливковый, темно-оливковый, оливковый \\

khaki & haki, хаки, темный хаки \\

swamp & болотный, болото, болотно серый, хаки болотный \\

mint & mint, soft mint, мятный, цвет мятного сиропа \\

pistachio & фисташковый, светло фисташковый \\

cyan & циановый, бирюзовый \\

neon & neon, неоновый \\

unknown & қою, күлгін, көктем, camel, butter cream, sage, цвет теннисного мяча, цвет искусственного газона, свинья в Minecraft, between brown and yellow... \\

orange & оранжевый, light orange, цвет цедры апельсина \\

white & white \\

ivory & слоновая кость, слоновый, айвори \\

clay & глиняный \\

\bottomrule
\end{tabularx}
\end{table}

\section{Results and analysis}

\begin{figure*}[ht]
\centering
\begin{subfigure}{0.32\textwidth}
    \centering
    \includegraphics[width=\linewidth]{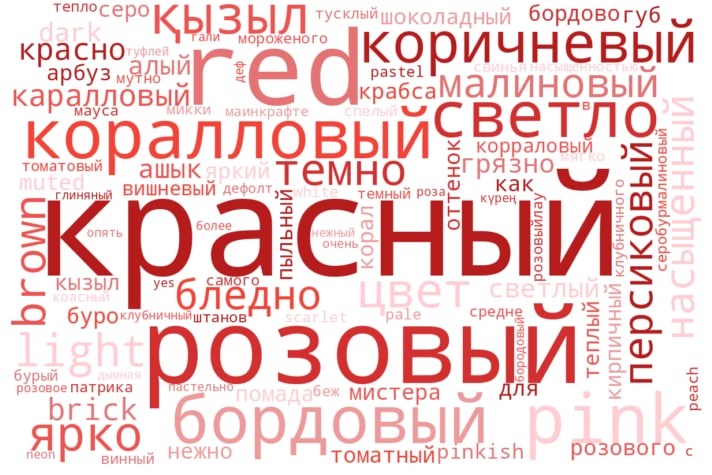}
    \caption{RED}
\end{subfigure}
\hfill
\begin{subfigure}{0.32\textwidth}
    \centering
    \includegraphics[width=\linewidth]{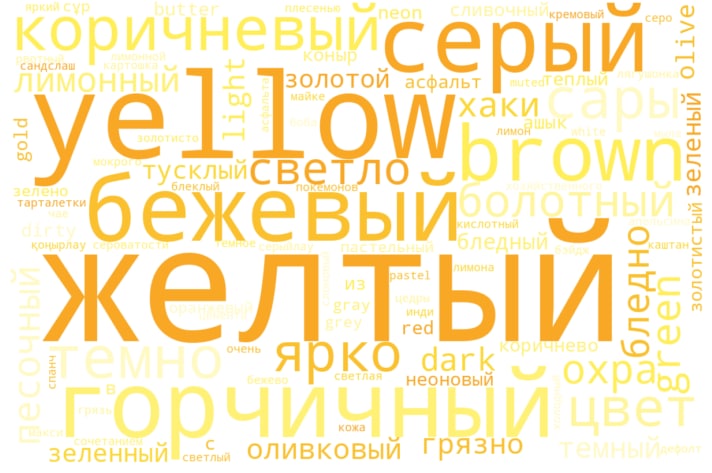}
    \caption{YELLOW}
\end{subfigure}
\hfill
\begin{subfigure}{0.32\textwidth}
    \centering
    \includegraphics[width=\linewidth]{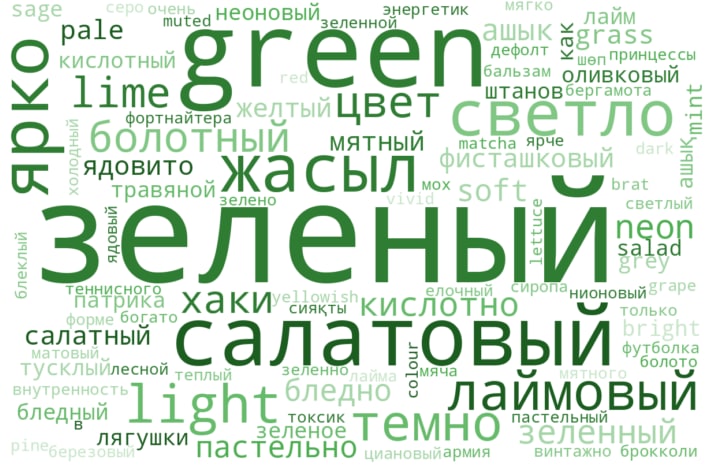}
    \caption{GREEN}
\end{subfigure}
\caption{Word clouds of color terms for RED, YELLOW, and GREEN colors}
\label{fig:word_clouds}
\end{figure*}


The word clouds in   Fig. \ref{fig:word_clouds} reveal different naming patterns for each hue. 
\textit{Green} responses were dominated by \textit{``green''}, \textit{``зеленый''}, and \textit{``жасыл''}, showing strong agreement among participants. 
\textit{Red} responses were mainly associated with \textit{``red''} and \textit{``красный''}, but also included many neighboring terms such as \textit{``pink''}, \textit{``coral''}, and \textit{``burgundy''}. 
Yellow showed the highest diversity, with frequent alternative names such as \textit{``brown'', ``mustard'', ``beige'', ``gray''}, and \textit{``golden''}.

Table~\ref{tab:color-naming-v1} confirms these differences quantitatively. \textit{Green} achieved the highest consistency score, with 65.55\% of responses mapped to the green category. \textit{Red} showed moderate consistency at 45.87\%. At the same time, a considerable number of responses were mapped to \textit{pink} (20.49\%), \textit{coral} (7.95\%), \textit{brown} (5.50\%), and \textit{burgundy} (4.89\%). This shows that \textit{red} shades partially overlap with adjacent color categories when the stimulus becomes lighter or less saturated. \textit{Yellow} had the lowest consistency score at 39.06\%. A large proportion of responses were distributed across \textit{brown} (9.06\%), \textit{unknown} (8.44\%), \textit{gray} (7.50\%), \textit{mustard} (6.88\%), and \textit{beige} (6.88\%).

\begin{table}[ht]
\centering
\caption{Color naming results by hue category}
\begin{tabularx}{\columnwidth}{Xcc|Xcc|Xcc}
\toprule
\multicolumn{3}{c|}{\textbf{Green}} & 
\multicolumn{3}{c|}{\textbf{Red}} & 
\multicolumn{3}{c}{\textbf{Yellow}} \\
\textbf{Category} & \textbf{N} & \textbf{\%} & 
\textbf{Category} & \textbf{N} & \textbf{\%} & 
\textbf{Category} & \textbf{N} & \textbf{\%} \\
\midrule

\textbf{green} & \textbf{215} & \textbf{65.55} & 
\textbf{red} & \textbf{150} & \textbf{45.87} & 
\textbf{yellow} & \textbf{125} & \textbf{39.06} \\

salad & 39 & 11.89 & pink & 67 & 20.49 & brown & 29 & 9.06 \\
lime & 19 & 5.79 & unknown & 35 & 10.70 & unknown & 27 & 8.44 \\
unknown & 18 & 5.49 & coral & 26 & 7.95 & gray & 24 & 7.50 \\
swamp & 8 & 2.44 & brown & 18 & 5.50 & beige & 22 & 6.88 \\
khaki & 7 & 2.13 & burgundy & 16 & 4.89 & mustard & 22 & 6.88 \\
mint & 6 & 1.83 & peach & 6 & 1.83 & green & 13 & 4.06 \\
pistachio & 3 & 0.91 & brick & 3 & 0.92 & swamp & 10 & 3.12 \\
neon & 3 & 0.91 & maline & 3 & 0.92 & golden & 8 & 2.50 \\
cyan & 2 & 0.61 & clay & 1 & 0.31 & lemon & 7 & 2.19 \\
gray & 2 & 0.61 & gray & 1 & 0.31 & ohra & 7 & 2.19 \\
herbs & 2 & 0.61 & white & 1 & 0.31 & khaki & 6 & 1.88 \\
olive & 2 & 0.61 &  &  &  & sand & 5 & 1.56 \\
red & 1 & 0.30 &  &  &  & orange & 4 & 1.25 \\
yellow & 1 & 0.30 &  &  &  & olive & 4 & 1.25 \\
 &  &  &  &  &  & ivory & 3 & 0.94 \\
 &  &  &  &  &  & red & 2 & 0.62 \\
 &  &  &  &  &  & clay & 1 & 0.31 \\
 &  &  &  &  &  & white & 1 & 0.31 \\

\bottomrule
\end{tabularx}
\label{tab:color-naming-v1}
\end{table}

\begin{figure}[tb]
    \centering
    \includegraphics[width=1\linewidth]{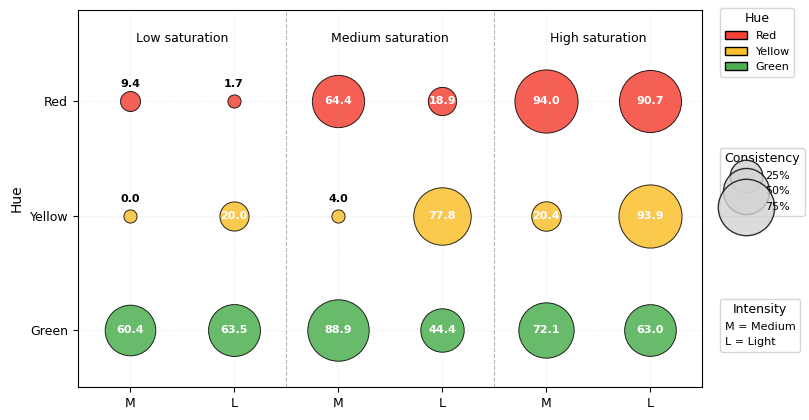}
    \caption{Consistency scores by hue, saturation, and intensity}
    \label{fig_hsi_cons}
\end{figure}


Next, Saturation clearly affected the consistency of color naming (see Fig.~\ref{fig_hsi_cons}). High-saturation stimuli were named most consistently (73.65\%), while low-saturation stimuli showed the lowest consistency (25.24\%) and the highest lexical diversity. As for Intensity, light stimuli were slightly more consistent than medium-intensity stimuli (53.88\% vs. 46.60\%).  At the Hue level, Saturation and Intensity affect the three hues differently.  \textit{Red} is most consistent under high saturation, especially in high-medium (94\%) and high-light (90.7\%) conditions, but becomes ambiguous at low saturation. 
\textit{Green} remains relatively stable, reaching its highest consistency in the medium-medium condition (88.9\%). 
\textit{Yellow} is the most sensitive to intensity: it is highly consistent under high-light (93.9\%) and medium-light (77.78\%) conditions, but drops sharply under medium intensity, especially in high-medium (20.41\%) and medium-medium (4.00\%).

Beyond direct color terms, participants often described colors through associations with foods, plants, and natural objects. As shown in Table~\ref{tab:semantic-categories}, food-related references constituted 16.82\% of all responses, comprising vegetable and plant-based descriptors (11.08\%) and fruit- and berry-related descriptors (5.74\%). Natural and earth-related associations accounted for an additional 8.92\% of responses. These findings indicate that participants frequently relied on familiar objects and experiences when naming colors.



\begin{table}[t]
\centering
\caption{Distribution of color terms by detected language}
\begin{tabularx}{\columnwidth}{XXXXXX}
\toprule
\textbf{Hue} & \textbf{en (\%)} & \textbf{kk (\%)} & \textbf{ru (\%)} & \textbf{mixed (\%)} & \textbf{Responses}\\
\midrule
green  & 22.87 & 7.01 & 69.82 & 0.3 & 328 \\
red    & 22.02 & 4.28 & 72.48 & 1.22 & 327 \\
yellow & 22.19 & 5.94 & 71.25 & 0.62 & 320 \\
\midrule
\textbf{Total} & \textbf{218} & \textbf{56} & \textbf{694} & \textbf{7} & 975 \\
\bottomrule
\end{tabularx}
\label{tab:language-distribution}
\end{table}








\begin{table}[t]
\caption{Semantic categories identified in free color naming responses}
\label{tab:semantic-categories}
\centering
\footnotesize
\begin{tabularx}{\columnwidth}{lXcc}
\toprule
\textbf{Category} & \textbf{Representative examples} & \textbf{Count} & \textbf{(\%)} \\
\midrule

basic color &
red, green, yellow, красный, жасыл &
350 & 35.90 \\

shade color &
light green, dark red, бледно-желтый, ярко зеленый, muted yellow &
310 & 31.79 \\

food, fruit, berry &
lemon, lime, peach, арбуз, малиновый &
56 & 5.74 \\

vegetable, plant &
salad, lettuce, брокколи, мятный, фисташковый &
108 & 11.08 \\

earth, natural &
brick, concrete, болотный, песочный, охра &
87 & 8.92 \\

unclear, other &
between brown and yellow, көктем, color of a tennis ball &
64 & 6.56 \\

\bottomrule
\end{tabularx}
\end{table}

\begin{figure}[ht]
\centering

\begin{subfigure}[b]{\columnwidth}
    \centering
    \includegraphics[width=0.73\columnwidth]{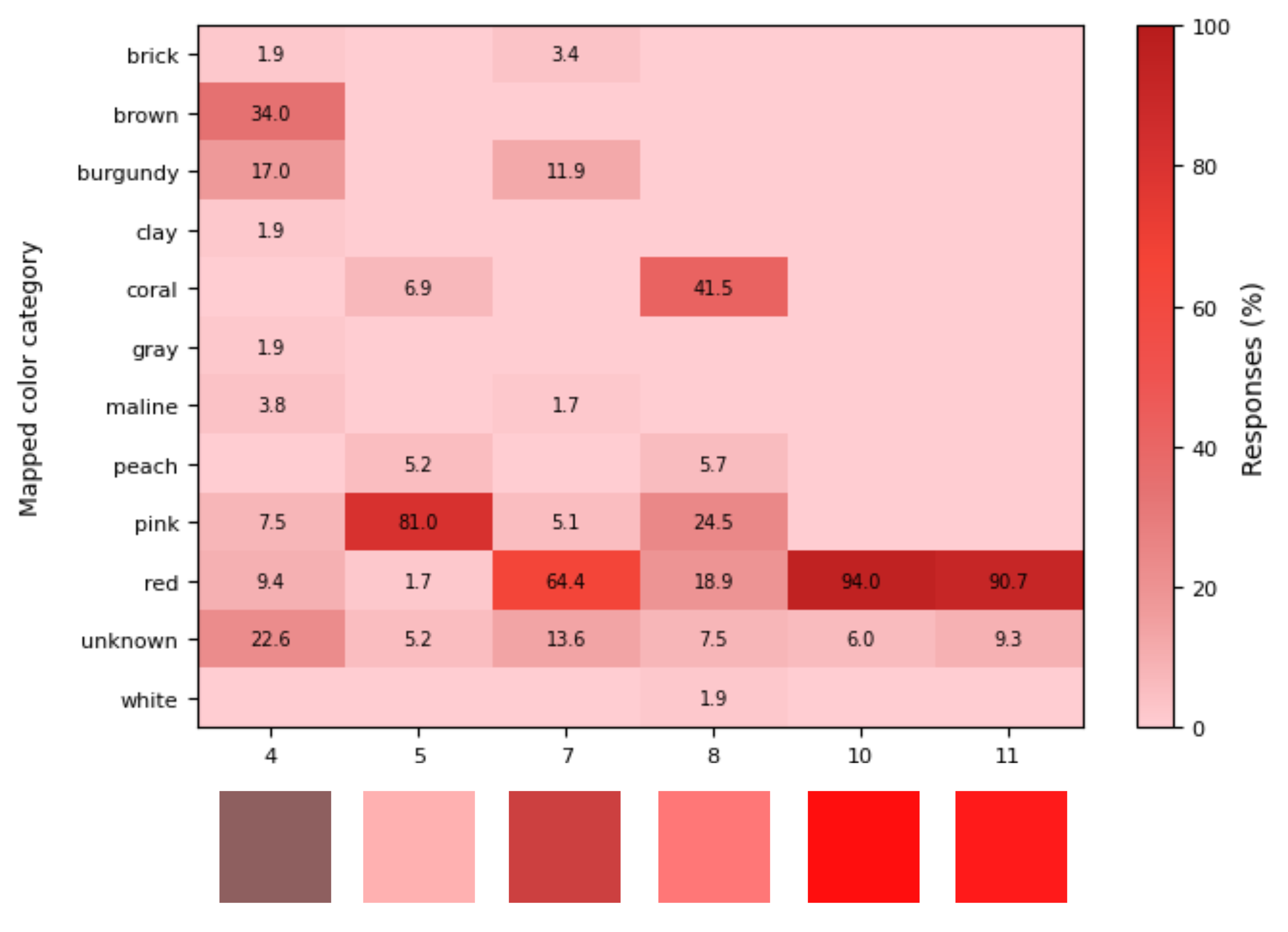}
    \caption{RED}
    \label{fig:heatmap-red}
\end{subfigure}

\vspace{0.1cm}

\begin{subfigure}[b]{\columnwidth}
    \centering
    \includegraphics[width=0.73\columnwidth]{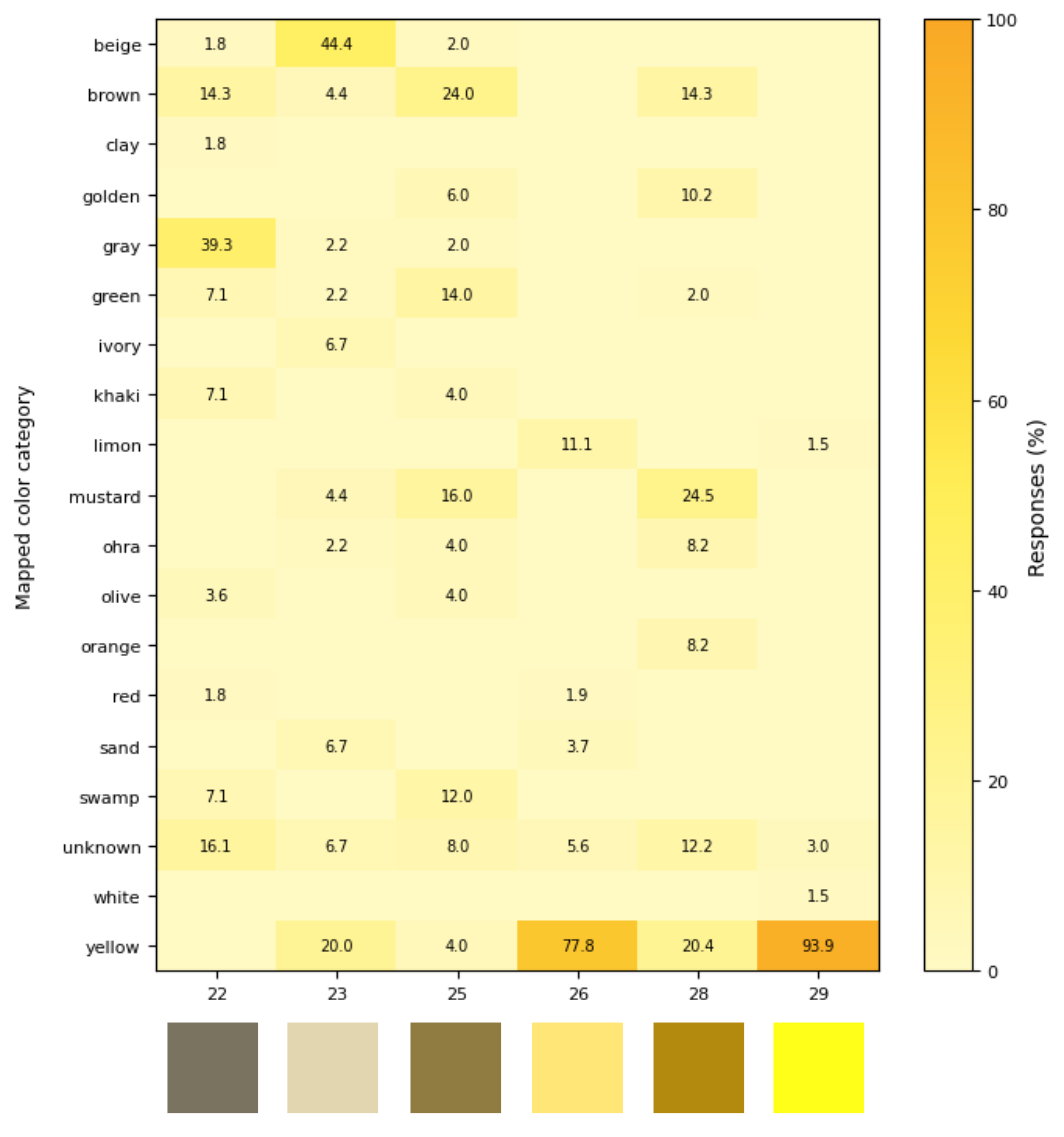}
    \caption{YELLOW}
    \label{fig:heatmap-yellow}
\end{subfigure}

\vspace{0.1cm}

\begin{subfigure}[b]{\columnwidth}
    \centering
    \includegraphics[width=0.73\columnwidth]{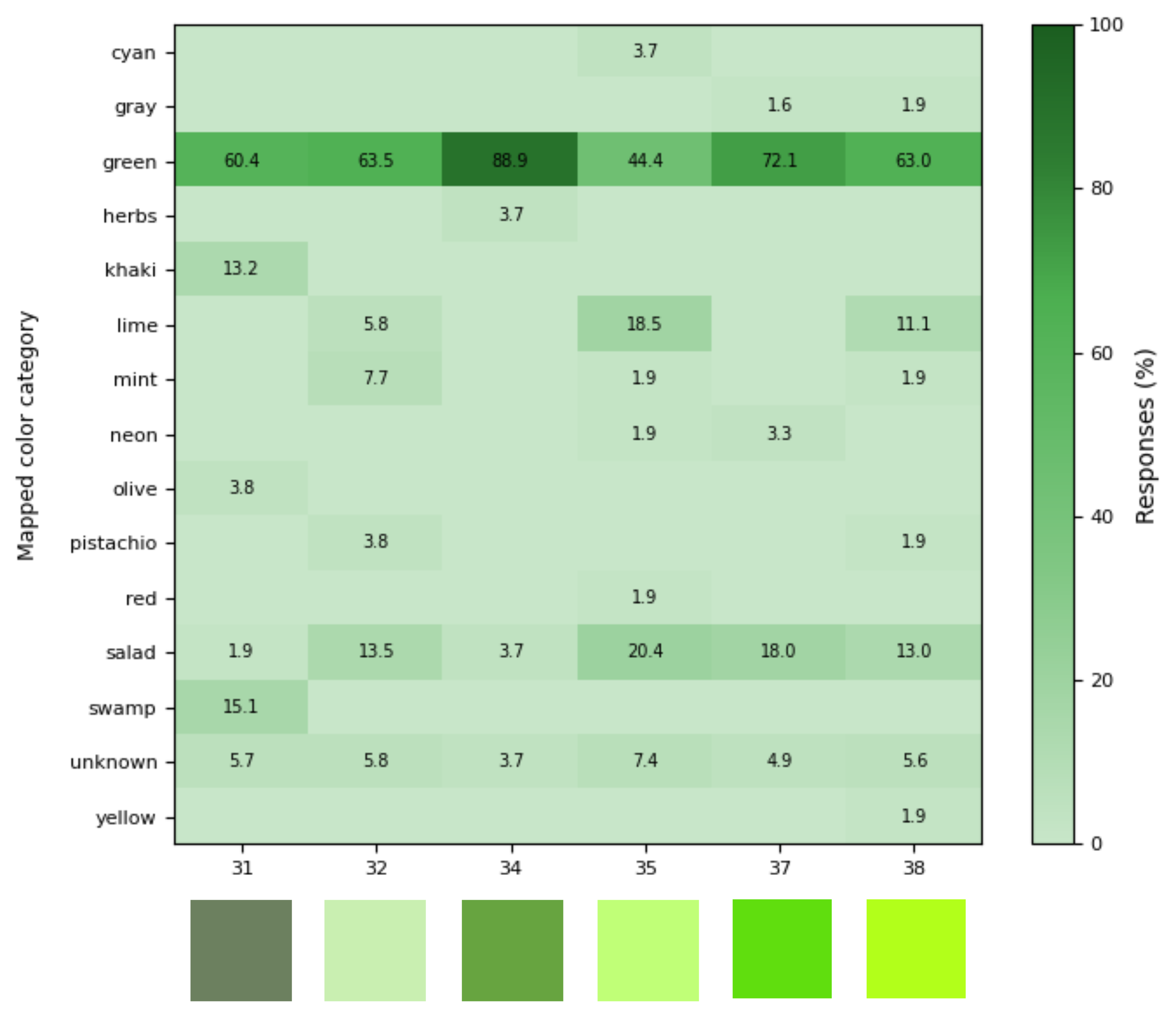}
    \caption{GREEN}
    \label{fig:heatmap-green}
\end{subfigure}

\caption{Percentage heatmap of mapped color categories for RED, YELLOW, and GREEN hues}
\label{fig:heatmaps}
\end{figure}


The heatmap analysis in Fig.~\ref{fig:heatmaps} shows that \textit{Green} is the most stable hue category. Most \textit{Green} stimuli were mapped to the green category at high percentages, especially ID 34 (88.9\%) and ID 37 (72.1\%), whereas only ID 35 showed greater ambiguity among green, salad, and lime. \textit{Red} shows moderate consistency (in Fig.~\ref{fig:heatmap-red}) and depends strongly on shade. Some \textit{red} stimuli were clearly mapped to red, such as ID 10 (94.0\%) and ID 11 (90.7\%), whereas other samples shifted toward neighboring categories, including pink for ID 5 and coral for ID 8. In contrast, \textit{Yellow} (Fig.~\ref{fig:heatmap-yellow}) is the most ambiguous hue category. Although ID 29 (93.9\%) and ID 26 (77.8\%) were consistently mapped to yellow, other \textit{yellow} stimuli were frequently perceived as brown, gray, mustard, beige, or unknown, indicating high sensitivity to changes in saturation and intensity.

As for gender-related differences, the semantic analysis showed that female participants used basic color terms and shade-based descriptions more frequently, whereas male participants used more associative descriptions related to fruits, berries, vegetables, and plants (female: 13.04\% vs. male: 19.75\%). 
Overall consistency was similar for female and male respondents, with female participants showing 51.21\% consistency and male participants showing 49.64\%. These findings suggest that gender does not strongly affect overall color naming consistency.

Overall, the results show the color categories exhibit unequal levels of perceptual stability. Color-naming consistency follows the order: \textit{Green} (65.55\%)> \textit{Red} (45.87\%)> \textit{Yellow} (39.06\%). \textit{Green} shades were consistently identified as green despite noticeable variations in appearance. It suggests that \textit{Green} is a relatively broad perceptual category. In contrast, \textit{Yellow} shades generated a wider range of alternative names, including \textit{gold, beige, olive, mustard, and brown-like descriptions}, indicating narrower category boundaries and greater naming ambiguity. The \textit{Red} shades demonstrated intermediate behavior between these two extremes.



\section{Discussion}





From the analysis of the collected responses, we observed that participants frequently described colors by referencing familiar natural and everyday objects. In particular, responses included names of fruits and vegetables (e.g., \textit{"tomato", "pistachio", "peach"}) (see  Table~\ref{tab:semantic-categories}). This phenomenon is consistent with findings in cognitive linguistics and color-naming research, where color categorization is often grounded in culturally salient prototypes rather than in abstract perceptual dimensions \cite{berlin1969basic, regier2007}. The use of object-based descriptors (e.g., fruit and food) has also been reported in studies of embodied cognition and the use of perceptual metaphor \cite{lakoff1980metaphors, barsalou2008grounded}.

Studies of English-speaking participants have shown that women tend to produce a richer color vocabulary and use non-basic color terms more frequently than men, while men rely more heavily on basic color terms~\cite{mylonas2010, paramei2018russian}. 
Female participants showed a preference for object-derived terms such as \textit{persikovyj} (peach) and 
\textit{bezevyj} (beige), whereas male participants more frequently used basic terms and descriptors associated with natural colors~\cite{paramei2018russian}. 
Although prior research has consistently documented gender-related differences in color naming behavior, naming consistency in our study did not differ significantly between genders.

\section{Conclusion}
This paper investigates the consistency of \textit{Red, Yellow}, and \textit{Green} color categories through a free color-naming experiment using 18 color samples from the COLIBRI dataset. The results show that color categories differ in their naming consistency and perceptual stability. In particular, \textit{Green} shades were generally identified more consistently, while \textit{Yellow} shades were associated with a wider variety of color names. These findings suggest that color categories occupy different perceptual regions and exhibit varying degrees of stability to changes in color appearance. Future work will extend the analysis to additional color categories.

\section*{Acknowledgment}
We express gratitude to all participants who took part in the free color-naming experiment conducted during the 4th Annual Graduate Student Research Workshop on Computer Science, Information Systems, and Engineering at KBTU. This work was supported by the Science Committee of the Ministry of Science and Higher Education of the Republic of Kazakhstan (Grant No. AP22786412)


\bibliography{bibfile}

\begin{thebibliography}{10}
\providecommand{\url}[1]{#1}
\csname url@samestyle\endcsname
\providecommand{\newblock}{\relax}
\providecommand{\bibinfo}[2]{#2}
\providecommand{\BIBentrySTDinterwordspacing}{\spaceskip=0pt\relax}
\providecommand{\BIBentryALTinterwordstretchfactor}{4}
\providecommand{\BIBentryALTinterwordspacing}{\spaceskip=\fontdimen2\font plus
\BIBentryALTinterwordstretchfactor\fontdimen3\font minus \fontdimen4\font\relax}
\providecommand{\BIBforeignlanguage}[2]{{%
\expandafter\ifx\csname l@#1\endcsname\relax
\typeout{** WARNING: IEEEtran.bst: No hyphenation pattern has been}%
\typeout{** loaded for the language `#1'. Using the pattern for}%
\typeout{** the default language instead.}%
\else
\language=\csname l@#1\endcsname
\fi
#2}}
\providecommand{\BIBdecl}{\relax}
\BIBdecl

\bibitem{Twomey2021}
\BIBentryALTinterwordspacing
C.~R. Twomey, G.~Roberts, D.~H. Brainard, and J.~B. Plotkin, ``What we talk about when we talk about colors,'' \emph{Proceedings of the National Academy of Sciences}, vol. 118, no.~39, 2021. [Online]. Available: \url{http://dx.doi.org/10.1073/pnas.2109237118}
\BIBentrySTDinterwordspacing

\bibitem{shamoi2025colibrifuzzymodelcolor}
P.~Shamoi, N.~Toganas, M.~Muratbekova, E.~Kadyrgali, A.~Yerkin, A.~Igali, M.~Ziyada, A.~Adilova, A.~Karatayev, and Y.~Torekhan, ``Colibri fuzzy model: Color linguistic-based representation and interpretation,'' \emph{IEEE Access}, vol.~13, pp. 205\,932--205\,956, 2025.

\bibitem{Chamorro-Martínez2017Fuzzy}
J.~Chamorro-Martínez, J.~M. Soto-Hidalgo, P.~Martínez-Jiménez, and D.~Sánchez, ``Fuzzy color spaces: A conceptual approach to color vision,'' \emph{IEEE Transactions on Fuzzy Systems}, vol.~25, pp. 1264--1280, 2017.

\bibitem{Witzel2019Variation}
C.~Witzel, ``Variation of saturation across hue affects unique and typical hue choices,'' \emph{i-Perception}, vol.~10, 2019.

\bibitem{Witzel2013Focal}
C.~Witzel, J.~Maule, and A.~Franklin, ``Focal colors as perceptual anchors of color categories,'' \emph{Journal of Vision}, vol.~13, pp. 1164--1164, 2013.

\bibitem{berlin1969basic}
B.~Berlin and P.~Kay, \emph{Basic Color Terms: Their Universality and Evolution}.\hskip 1em plus 0.5em minus 0.4em\relax University of California Press, 1969.

\bibitem{mylonas2020}
D.~Mylonas and L.~D. Griffin, ``Coherence of achromatic, primary and basic classes of colour categories,'' \emph{Vision Research}, vol. 175, pp. 14--22, 2020.

\bibitem{Lindsey2014The}
D.~Lindsey and A.~M. Brown, ``The color lexicon of american english.'' \emph{Journal of vision}, vol. 14 2, 2014.

\bibitem{Lindsey2006Universality}
D.~T. Lindsey and A.~M. Brown, ``Universality of color names,'' \emph{Proceedings of the National Academy of Sciences}, vol. 103, pp. 16\,608--16\,613, 2006.

\bibitem{kay1997color}
P.~Kay and B.~Berlin, ``Color naming across languages,'' \emph{Language}, 1997.

\bibitem{winawer2007color}
J.~e.~a. Winawer, ``Russian blues reveal effects of language on color discrimination,'' \emph{PNAS}, 2007.

\bibitem{sabitkyzy2026universalcolornamingsystem}
\BIBentryALTinterwordspacing
A.~Sabitkyzy, M.~Shagyrov, and P.~Shamoi, ``Toward a universal color naming system: A clustering-based approach using multisource data,'' 2026. [Online]. Available: \url{https://arxiv.org/abs/2604.03235}
\BIBentrySTDinterwordspacing

\bibitem{witzel2018}
C.~Witzel and K.~R. Gegenfurtner, ``Color perception: Objects, constancy, and categories,'' \emph{Annual Review of Vision Science}, vol.~4, pp. 475--499, 2018.

\bibitem{regier2007}
T.~Regier, P.~Kay, and N.~Khetarpal, ``Color naming reflects optimal partitions of color space,'' \emph{Proceedings of the National Academy of Sciences}, vol. 104, no.~4, pp. 1436--1441, 2007.

\bibitem{Shamoi14}
P.~Shamoi, A.~Inoue, and H.~Kawanaka, ``Fuzzy color space for apparel coordination,'' \emph{Open Journal of Information Systems (OJIS)}, vol.~1, pp. 20--28, 01 2014.

\bibitem{Regier2005Focal}
T.~Regier, P.~Kay, and R.~S. Cook, ``Focal colors are universal after all,'' \emph{Proceedings of the National Academy of Sciences}, vol. 102, no.~23, pp. 8386--8391, 2005.

\bibitem{Abbott2016Focal}
J.~T. Abbott, T.~L. Griffiths, and T.~Regier, ``Focal colors across languages are representative members of color categories,'' \emph{Proceedings of the National Academy of Sciences}, vol. 113, no.~40, pp. 11\,178--11\,183, 2016.

\bibitem{Morimoto2022Invariant}
T.~Morimoto, Y.~Yamauchi, and K.~Uchikawa, ``Invariant categorical color regions across illuminant change coincide with focal colors,'' \emph{Journal of Vision}, vol.~22, no.~14, p. 3208, 2022.

\bibitem{emery2017}
K.~J. Emery, V.~J. Volbrecht, D.~H. Peterzell, and M.~A. Webster, ``Variations in normal color vision. {VII}. relationships between color naming and hue scaling,'' \emph{Vision Research}, vol. 141, pp. 66--75, 2017.

\bibitem{cibelli2016}
E.~Cibelli, Y.~Xu, J.~L. Austerweil, T.~L. Griffiths, and T.~Regier, ``The {Sapir-Whorf} hypothesis and probabilistic inference: Evidence from the domain of color,'' \emph{PLoS ONE}, vol.~11, no.~7, p. e0158725, 2016.

\bibitem{gong2019}
T.~Gong, H.~Gao, Z.~Wang, and L.~Shuai, ``Perceptual constraints on colours induce the universality of linguistic colour categorisation,'' \emph{Scientific Reports}, vol.~9, p. 7719, 2019.

\bibitem{shunbo2025}
S.~Dai and A.~Z. Zainal, ``An empirical study of basic color terms in chinese,'' \emph{Forum for Linguistic Studies}, vol.~7, no.~5, pp. 861--872, 2025.

\bibitem{abdou2021}
M.~Abdou, A.~Kulmizev, D.~Hershcovich, S.~Frank, E.~Pavlick, and A.~S{\o}gaard, ``Can language models encode perceptual structure without grounding? {A} case study in color,'' in \emph{Proceedings of the 25th Conference on Computational Natural Language Learning ({CoNLL})}.\hskip 1em plus 0.5em minus 0.4em\relax Association for Computational Linguistics, 2021, pp. 109--132.

\bibitem{forder2017}
L.~Forder, X.~He, and A.~Franklin, ``Colour categories are reflected in sensory stages of colour perception when stimulus issues are resolved,'' \emph{PLoS ONE}, vol.~12, no.~5, p. e0178097, 2017.

\bibitem{burambekova25}
A.~Burambekova and P.~Shamoi, ``Comparative analysis of color models for human perception and visual color difference,'' in \emph{2025 IEEE 5th International Conference on Smart Information Systems and Technologies (SIST)}, 2025, pp. 1--6.

\bibitem{Color_model}
M.~Muratbekova, N.~Toganas, A.~Igali, M.~Shagyrov, E.~Kadyrgali, A.~Yerkin, and P.~Shamoi, ``Color models in image processing: a review and experimental comparison,'' \emph{Discover Applied Sciences}, vol.~8, no.~5, p. 494, Mar 2026.

\bibitem{lakoff1980metaphors}
G.~Lakoff and M.~Johnson, \emph{Metaphors We Live By}.\hskip 1em plus 0.5em minus 0.4em\relax University of Chicago Press, 1980.

\bibitem{barsalou2008grounded}
L.~W. Barsalou, ``Grounded cognition,'' \emph{Annual Review of Psychology}, 2008.

\bibitem{mylonas2010}
D.~Mylonas and L.~MacDonald, ``Online colour naming experiment using {Munsell} colour samples,'' in \emph{Proceedings of the 5th European Conference on Colour in Graphics, Imaging, and Vision (CGIV)}.\hskip 1em plus 0.5em minus 0.4em\relax Joensuu, Finland: IS\&T, 2010, pp. 27--32.

\bibitem{paramei2018russian}
G.~V. Paramei, Y.~A. Griber, and D.~Mylonas, ``An online color naming experiment in {Russian} using {Munsell} color samples,'' \emph{Color Research \& Application}, vol.~43, pp. 358--374, 2018.

\end{thebibliography}

\end{document}